\documentclass{article}
\usepackage{times}
\usepackage{graphicx} 
\usepackage{caption}
\usepackage{subcaption}
\usepackage{natbib}
\usepackage{amsmath}
\usepackage{amssymb}
\usepackage{amsthm}
\usepackage{amsfonts}
\usepackage{algorithm}
\usepackage{algorithmic}

\usepackage[nohyperref,accepted]{icml2016}

\usepackage{adjustbox}
\usepackage{array}
\usepackage[table,x11names]{xcolor}

\newcolumntype{R}[2]{%
    >{\adjustbox{angle=#1,lap=\width-(#2)}\bgroup}%
    l%
    <{\egroup}%
}
\newcommand*\rot{\multicolumn{1}{R{45}{1em}}}

\icmltitlerunning{Hierarchical Compound Poisson Factorization}

\newtheorem{defn}{Definition}
\newtheorem{theorem}{Theorem}
\newtheorem{remark}{Remark}

\begin{document}

\twocolumn[
\icmltitle{Hierarchical Compound Poisson Factorization}

\icmlauthor{Mehmet E. Basbug}{mehmetbasbug@yahoo.com}
\icmladdress{Princeton University,
             35 Olden St., Princeton, NJ 08540 USA}
\icmlauthor{Barbara E. Engelhardt}{bee@princeton.edu}
\icmladdress{Princeton University,
             35 Olden St., Princeton, NJ 08540 USA}
\icmlkeywords{compound poisson, factorization, variational inference}

\vskip 0.3in
]

\begin{abstract}
Non-negative matrix factorization models based on a hierarchical Gamma-Poisson structure capture user and item behavior effectively in extremely sparse data sets, making them the ideal choice for collaborative filtering applications. Hierarchical Poisson factorization (HPF) in particular has proved successful for scalable recommendation systems with extreme sparsity. HPF, however, suffers from a tight coupling of sparsity model (absence of a rating) and response model (the value of the rating), which limits the expressiveness of the latter. Here, we introduce hierarchical compound Poisson factorization (HCPF) that has the favorable Gamma-Poisson structure and scalability of HPF to high-dimensional extremely sparse matrices. More importantly, HCPF decouples the sparsity model from the response model, allowing us to choose the most suitable distribution for the response. HCPF can capture binary, non-negative discrete, non-negative continuous, and zero-inflated continuous responses. We compare HCPF with HPF on nine discrete and three continuous data sets and conclude that HCPF captures the relationship between sparsity and response better than HPF.
\end{abstract}

\section{Introduction}
Matrix factorization has been a central subject in statistics since the introduction of principal component analysis (PCA)~\cite{pearson1901liii}. The goal is to embed data into a lower dimensional space with minimal loss of information. The dimensionality reduction aspect of matrix factorization has become increasingly important in exploratory data analysis as the dimensionality of data has exploded.

One alternative to PCA, non-negative matrix factorization (NMF), was first developed for factorizing matrices for face recognition~\cite{lee1999learning}. The idea behind NMF is that the contributions of each feature to a factor are non-negative. Although the motivation behind this choice has roots in cerebral representations of objects, non-negativeness has found great appeal in applications such as collaborative filtering~\cite{gopalan2013scalable}, document classification~\cite{xu2003document}, and signal processing~\cite{fevotte2009nonnegative}.

In collaborative filtering, the data are highly sparse user by item response matrices. For example, the  \textit{Netflix} movie rating data set includes $480$K users, $17$K movies and $100$M ratings, meaning that $0.988$ of the matrix entries are missing. In the \textit{donors-choose} data set, the user (donor) response to an item (project) quantifies their monetary donation to that project; this matrix includes $1.3$M donors, $525$K projects and $2.66$M donations, meaning that $0.999996$ of the matrix entries are missing.

In the collaborative filtering literature, there are two school of thoughts on how to treat missing entries. The first one assumes that entries are missing at random; that is, we observe a uniformly sampled subset of the data~\cite{marlin2012collaborative}. Factorization is done on the premise that the response value provides all the information needed. The second method assumes that matrix entries are not missing at random, but instead there is a underlying mixture model: first, a coin is flipped to determine if an entry is missing. If the entry is missing, its value is set to zero; if it is not missing, the response is drawn from a specific distribution~\cite{marlin2009collaborative}. In this framework, we postulate that absence of an entry carries information about the item and the user, and this information can be exploited to improve the overall quality of the factorization. The difficult part is in representing the connection between the absence of a response (sparsity model) and the numerical value of a response (response model)~\cite{little2014statistical}. We are concerned with the problem of sparse matrix factorization where the data are not missing at random.

Extensions to the NMF hint at a model that addresses this problem. Recent work showed that the NMF objective function is equivalent to a factorized Poisson likelihood~\cite{cemgil2009bayesian}. The authors proposed a Bayesian treatment of the Poisson model with Gamma conjugate priors on the latent factors, laying the foundation for hierarchical Poisson factorization (HPF)~\cite{gopalan2013scalable}. The Gamma-Poisson structure is also used in earlier work for matrix factorization because of its favorable behavior~\cite{canny2004gap,ma2011probabilistic}. Long tailed Gamma priors were found to be powerful in capturing the underlying user and item behavior in collaborative filtering problems by applying strong shrinkage to the values near zero but allowing the non-zero responses to escape shrinkage~\cite{polson2010shrink}. An extension of the Poisson factorization to non-discrete data using data augmentation has been considered~\cite{fevotte2009nonnegative,fevotte2011algorithms}. Along the similar lines, the connection between beta divergences and compound Poisson Gamma distribution is exploited to develop a non-negative matrix factorization model for sparse positive data~\cite{simsekli2013learning}. More recent work introduced a stochastic variational inference algorithm for scalable collaborative filtering using HPF~\cite{gopalan2013scalable}.

HPF models each factor contribution to be drawn from a Poisson distribution with a long tail gamma prior. Thanks to the additive property of Poisson, the sum of these contributions are again a Poisson random variable which is used to model the response. For a collaborative filtering problem, HPF treats missing entries as true zero responses when applied to both missing and non-missing entries~\cite{gopalan2013scalable}. When the overwhelming majority of the matrix is missing, the posterior estimates for the Poisson factors are close to zero. This mechanism has a profound impact on the response model: When the Poisson parameter of a zero truncated Poisson (ZTP) distribution approaches zero, the ZTP converges to a degenerate distribution at $1$. In other words, if we condition on the fact that an entry is not missing, the HPF predicts that the response value is $1$ with a very high probability. Since the response model and sparsity model are so tightly coupled, HPF is suitable only for sparse binary matrices. When using HPF on the full matrix (i.e., missing data and responses together), one might binarize the data to improve performance of the HPF~\cite{gopalan2013scalable}. However, binarization ignores the impact of the response model on absence. For instance, a user is more likely to watch a movie that is similar to a movie that she gave a high rating to relative to one that she rated lower. This information is ignored in the HPF model.

In this paper, we introduce hierarchical compound Poisson factorization (HCPF), which has the same Gamma-Poisson structure as the HPF model and is equally computationally tractable. HCPF differs from the HPC in that it flexibility decouples the sparsity model from the response model, allowing the HCPF to accurately model binary, non-negative discrete, non-negative continuous and zero-inflated continuous responses in the context of extreme sparsity. Unlike HPF, the ZTP distribution does not concentrate around $1$, but instead converges to the response distribution that we choose. In other words, we effectively decouple the sparsity model and the response model. Decoupling does not imply independence, but instead the ability to capture the distributional characteristics of the response more accurately in the presence of extreme sparsity. HCPF still retains the useful property of HPF that the expected non-missing response value is related to the probability of non-absence, allowing the sparsity model to exploit information from the responses.

First we generalize HPF to handle non-discrete data. In Section~\ref{sect:edm}, we introduce additive exponential dispersion models, a class of probability distributions. We show that any member of the additive exponential dispersion model family, including normal, gamma, inverse Gaussian, Poisson, binomial, negative binomial, and zero-truncated Poisson, can be used for the response model. In Section~\ref{sect:comp}, we prove that a compound Poisson distribution converges to its element additive EDM distribution as sparsity increases.

Section~\ref{sect:hcpf} describes the generative model for hierarchical compound Poisson factorization (HCPF) and the mean field stochastic variational inference (SVI) algorithm for HCPF, which allows us to fit HCPF to data sets with millions of rows and columns quickly~\cite{gopalan2013scalable}.  In Section~\ref{sect:results}, we show the favorable behavior of HCPF as compared to HPF on twelve data sets of varying size including ratings data sets (\textit{amazon, movielens, netflix} and \textit{yelp}), social media activity data sets (\textit{wordpress} and \textit{tencent}), a web activity data set (\textit{bestbuy}), a music data set (\textit{echonest}), a biochemistry data set (\textit{merck}), financial data sets (\textit{donation} and \textit{donorschoose}), and a genomics data set (\textit{geuvadis}).

\section{Exponential Dispersion Models}
\label{sect:edm}
Additive exponential dispersion models (EDMs) are a generalization of the natural exponential family where the nonzero dispersion parameter scales the log-partition function~\cite{jorgensen1997theory}. We first give a formal definition of additive EDM, and present seven useful members (Table~\ref{table:edm}).
\begin{table*}[t]
\caption{{\bf Seven common additive exponential dispersion models.} Normal, gamma, inverse Gaussian, Poisson, binomial, negative binomial, and zero truncated Poisson (ZTP) distributions written in additive EDM form with the variational distribution of the Poisson variable for the corresponding compound Poisson additive EDM. The gamma distribution is parametrized with shape ($a$) and rate ($b$).}\label{table:edm}
\vskip -0.65in
\begin{center}
\begin{small}
\begin{sc}
\begin{tabular}{llccccc}
\hline
\abovespace\belowspace
distribution & {} & $\theta$ & $\kappa$ & $\Psi(\theta)$ & $h(x,\kappa)$ & $q(n_{ui} = n)\propto$ \\
\hline
\abovespace
Normal & $N(\mu,\sigma^2)$ & $\frac{\mu}{\sigma^{2}}$ & $\sigma^2$ & $\frac{\theta^2}{2}$ & $\frac{1}{\sqrt{2\pi \kappa}}\exp(-\frac{x^2}{2\kappa})$ & $\exp\left \{ -\frac{n^2\mu^{2} + y_{ui}^{2}}{2n\sigma^{2}} \right \}\frac{\Lambda_{ui}^{n}}{n!\sqrt{n}}$\\
\belowspace
\abovespace
Gamma & $Ga(a,b)$ & $-b$ & $a$ & $-\log(-\theta)$ & $x^{\kappa-1}/\Gamma(\kappa)$ & $\frac{ (b^{a} y_{ui}^{a}\Lambda_{ui})^{n}}{\Gamma(n a) n!}$\\
\belowspace
\abovespace
Inv. Gaussian & $IG(\mu,\lambda)$ & $\frac{-\lambda}{2\mu^{2}}$ & $\sqrt{\lambda}$ & $-\sqrt{-2\theta}$ & $\frac{\kappa}{\sqrt{2\pi x^{3}}}\exp(-\frac{\kappa^2}{2x})$ & $\exp\left \{n \frac{\lambda}{\mu} - \frac{n^2 \lambda}{2y_{ui}} \right \}\frac{\Lambda_{ui}^{n}}{(n-1)!}$\\
\belowspace
\abovespace
Poisson & $Po(\lambda)$ & $\log \lambda$ & $1$ & $e^{\theta}$ & $\frac{\kappa^{x}}{x!}$ & $\exp\left \{-n \lambda \right \}\frac{n^{y_{ui}}\Lambda_{ui}^{n}}{n!}$\\
\belowspace
\abovespace
Binomial & $Bi(r,p)$ & $\log\frac{p}{1-p}$ & $r$ & $\log(1+e^\theta)$ & $\binom{\kappa}{x}$ & $\frac{(nr)!(1-p)^{nr}\Lambda_{ui}^{n}}{n! (nr-y_{ui})! }$\\
\belowspace
\abovespace
Neg. Binomial & $NB(r,p)$ & $\log p$ & $r$ & $-\log(1-e^\theta)$ & $\binom{x + \kappa -1}{x}$ & $\frac{(y_{ui} + nr -1)! (1-p)^{nr}\Lambda_{ui}^{n}}{n! (nr-1)!}$\\
\belowspace
\abovespace
ZTP & $ZTP(\lambda)$ & $\log \lambda$ & $1$ & $\log(e^{e^{\theta}} -1)$ & $\frac{1}{x!}\sum_{j=0}^{\kappa}(-1)^j (\kappa-j)^x \binom{\kappa}{j}$ & $\left ( \frac{\Lambda_{ui}}{e^{\lambda}-1}\right)^{n} \sum_{j=0}^{n} \frac{(-1)^j(n-j)^{y_{ui}}}{j!(n-j)!}$\\
\hline
\end{tabular}
\end{sc}
\end{small}
\end{center}
\vskip -0.25in
\end{table*}

\begin{defn}
A family of distributions $\mathcal{F}_{\Psi}= \left \{ p_{(\Psi,\theta,\kappa)} \mid \theta \in \Theta  = dom(\Psi) \subseteq \mathbb{R}, \kappa \in \mathbb{R}_{++} \right \}$ is called an \emph{additive exponential dispersion model} if
\begin{align}
p_{(\Psi,\theta,\kappa)}(x) &= \exp( x\theta - \kappa\Psi(\theta))h(x,\kappa)
\label{density_x}
\end{align}
where $\theta$ is the natural parameter, $\kappa$ is the dispersion parameter, $\Psi(\theta)$ is the base log-partition function, and $h(x,\kappa)$ is the base measure.
\end{defn}

The sum of additive EDMs with a shared natural parameter and base log partition function is again an additive EDM of the same type.
\begin{theorem}[Jorgensen, 1997]
Let $X_{1} \dots X_{M}$ be a sequence of additive EDMs such that $X_{i} \sim p_{\Psi}(x;\theta,\kappa_{i})$, then $X_{+} = X_{1} + \dots + X_{M} \sim p_{\Psi}(x;\theta,\sum_{i}\kappa_{i})$.
\label{th:edm}
\end{theorem}

In Table~\ref{table:edm}, we use a generalized definition of the zero truncated Poisson (ZTP) where the density of the sum of $\kappa$ i.i.d. random variables from an ordinary ZTP distribution can be expressed with the same formula~\cite{springael2006sum}. The last column in Table~\ref{table:edm} shows the variational update for the Poisson parameter of the compound Poisson additive EDM, further discussed in Section~\ref{sect:comp}.

\section{Compound Poisson Distributions}
\label{sect:comp}
\begin{figure*}[t]
    \begin{center}
      \begin{subfigure}[h]{0.32\textwidth}
        \includegraphics[width=\textwidth]{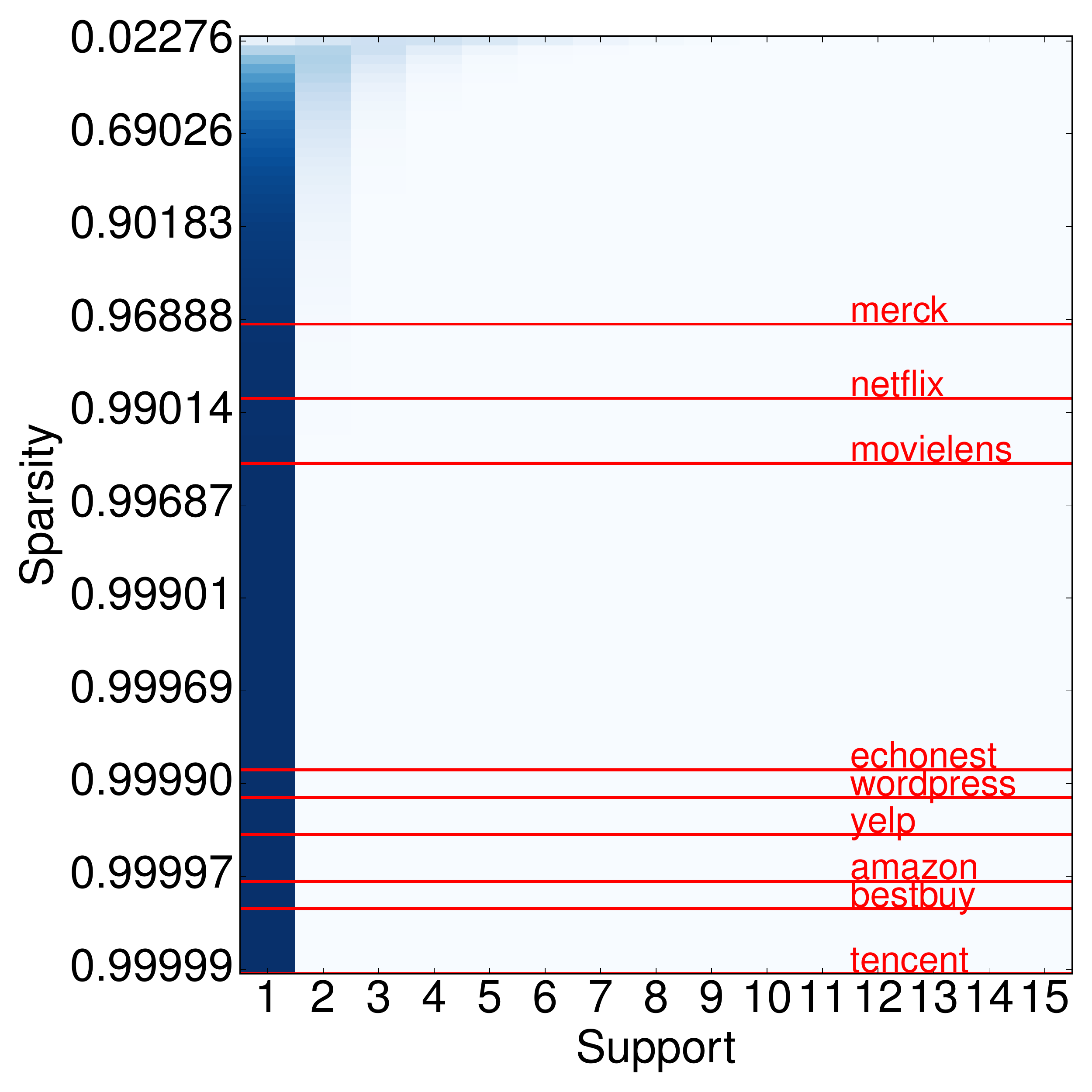}
        \caption{}
        \label{fig:xpp_pmf_0}
      \end{subfigure}
      ~
      \begin{subfigure}[h]{0.32\textwidth}
        \includegraphics[width=\textwidth]{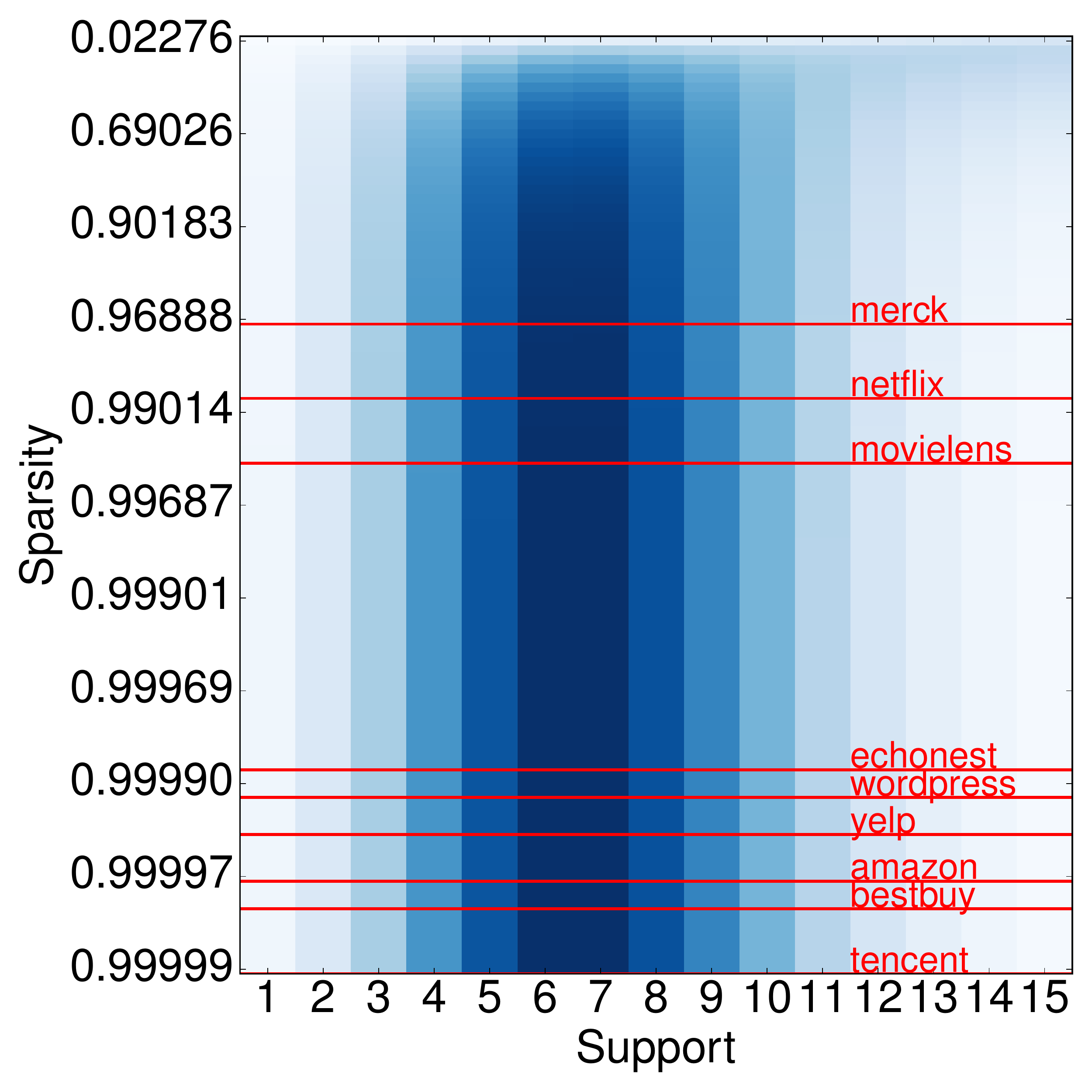}
        \caption{}
        \label{fig:xpp_pmf_7}
      \end{subfigure}
      ~
      \begin{subfigure}[h]{0.32\textwidth}
        \includegraphics[width=\textwidth]{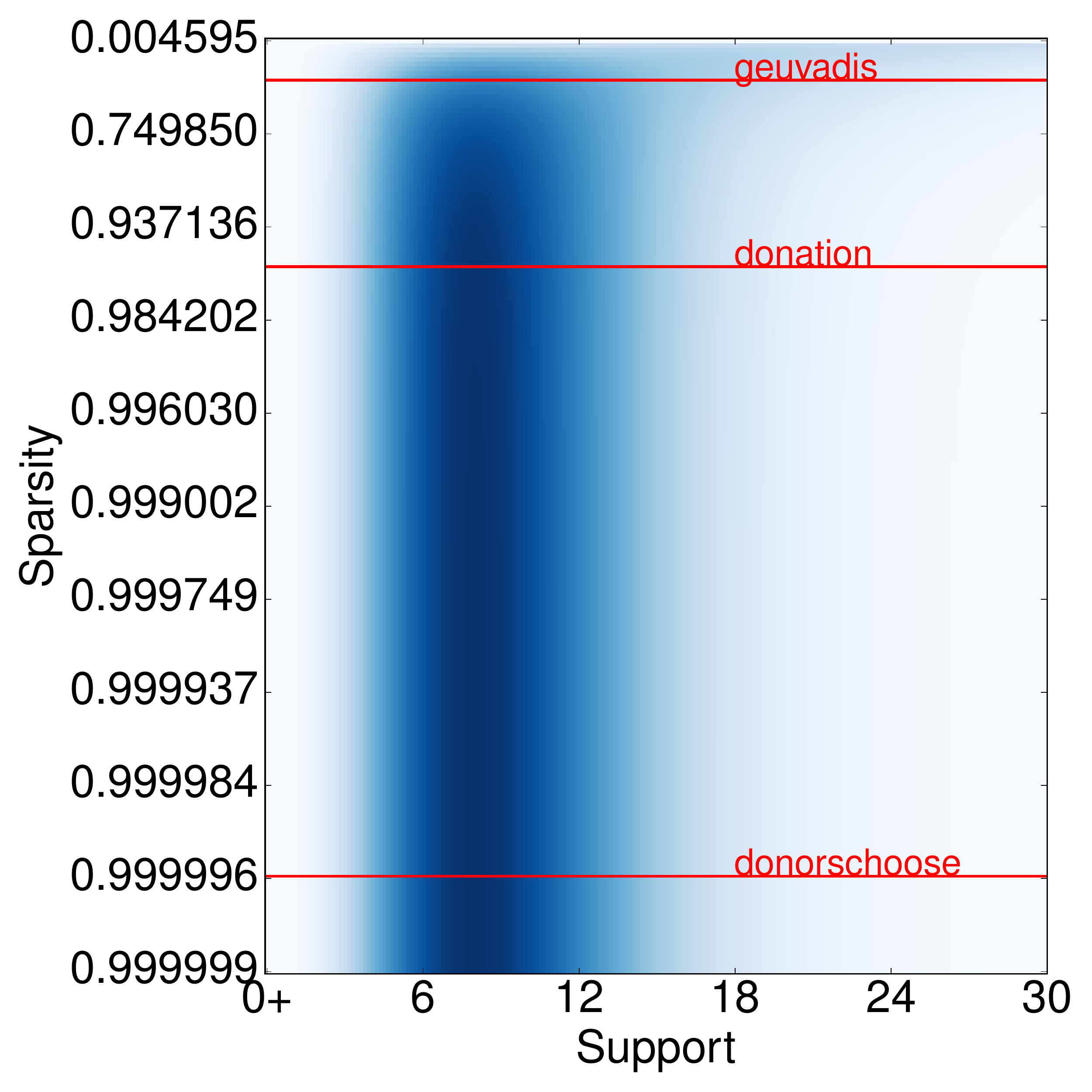}
        \caption{}
        \label{fig:xpp_pdf_gamma}
      \end{subfigure}
    \caption{{\bf PDF of the zero truncated compound Poisson random variable $X_{++}$ at various sparsity levels on log scale.} The PDF is color coded, where darker colors correspond to greater density. The response distribution is a) a degenerate $\delta_{1}$, b) a zero truncated Poisson with $\lambda = 7$, c) a gamma distribution with $a=5, b=0.5$. Red vertical lines mark the sparsity levels of various data sets.}
    \label{fig:xpp_pdf}
  \end{center}
  \vspace{-12pt}
\end{figure*}

We start with the general definition of a compound Poisson distribution and discuss compound Poisson additive EDM distributions. We then present the decoupling theorem and explain the implications of the theorem in detail.
\begin{defn}
Let $N$ be a Poisson distributed random variable with parameter $\Lambda$, $X_{1} ,\dots ,X_{N}$ be i.i.d. random variables distributed with an \emph{element distribution} $p_{\Psi}(x;\theta,\kappa)$. Then $X_{+} \doteq X_{1} + \dots + X_{N} \sim p_{\Psi}(x;\theta,\kappa,\Lambda)$ is called a \emph{compound Poisson random variable}.
\end{defn}

In general, $p_{\Psi}(x;\theta,\kappa,\Lambda)$ does not have a closed form expression, but it is a well defined probability distribution. The conditional form, $X_{+}\mid N$, is usually easier to manipulate, and we can calculate the marginal distribution of $X_{+}$ by integrating out $N$. When $N=0$, $X_{+}$ has a degenerate distribution at zero. Furthermore, if the element distribution is an additive EDM, we have the following theorem:
\begin{theorem}
\label{th:comp_edm}
Let $X \sim p_{\Psi}(x;\theta,\kappa)$ be an additive EDM and $X_{+} \sim p_{\Psi}(x;\theta,\kappa,\Lambda)$ be the compound Poisson random variable with the element random variable $X$, then
\begin{align}
X_{+} \mid N &= n \sim p_{\Psi}(x;\theta,n\kappa)\\
X_{+} \mid N &= 0 \sim \delta_{0}.
\end{align}
\end{theorem}

Theorem~\ref{th:comp_edm} implies that the conditional distribution of a compound Poisson additive EDM is again an additive EDM. Hence, both the conditional and the marginal densities of $X_{+}$ can be calculated easily.

We make the following remark before the decoupling theorem. Later, we show that HPF is a degenerate form of our model using this remark.
\begin{remark}
\label{th:degenerate}
Compound Poisson random variable $X_{+}$ is an ordinary Poisson random variable with parameter $\Lambda$ if and only if the element distribution is a degenerate distribution at $1$.
\end{remark}

We now present the decoupling theorem. This theorem shows that the distribution of a zero truncated compound Poisson random variable converges to its element distribution as the Poisson parameter ($\Lambda$) goes to zero.
\begin{theorem}
Let $X_{++} \doteq X_{+} \mid X_{+} \neq 0$ be a zero truncated compound Poisson random variable with element random variable $X$. If zero is not in the support of $X$, then
\begin{align}
Pr(X_{+} &= 0) = e^{-\Lambda}\label{eq:sparsity}\\
E[X_{++}] &= \frac{\Lambda}{1-e^{-\Lambda}} E[X]\label{eq:xpp_mean}\\
X_{++} &\overset{D}{\rightarrow} X \;\;\text{as} \;\;\Lambda\rightarrow 0.
\end{align}\label{th:convergence}
\end{theorem}
Proofs of Theorem~\ref{th:comp_edm},~\ref{th:convergence} and Remark~\ref{th:degenerate} can be found in the Appendix.

Let $X_{+}$ be a compound Poisson variable with element random distribution $p_{\Psi}(x;\theta,\kappa)$ and let $X_{++}$ be the zero truncated $X_{+}$ as in Theorem~\ref{th:convergence}. We can study the probability density function (PDF) of $X_{++}$ at various sparsity levels (Fig~\ref{fig:xpp_pdf}; Eq~\ref{eq:sparsity}) with respect to the average sparsity levels of our $9$ discrete data sets and $3$ continuous data sets. Importantly, the element distribution is nearly identical across all levels of sparsity.

We will use $X_{+}$ to model an entry of a full sparse matrix, meaning that we are including both missing and non-missing entries. The zero truncated random variable, $X_{++}$, corresponds to the non-missing response. In Fig~\ref{fig:xpp_pmf_0}, $X_{+}$ is an ordinary Poisson variable. As Remark~\ref{th:degenerate} and Theorem~\ref{th:convergence} suggest, at levels of extreme sparsity (i.e., $>90\%$ zeros), almost all of the probability mass of $X_{++}$ concentrates at $1$. That is, HPF predicts that, if an entry is not missing, then its value is $1$ with a high probability. To get a more flexible response model, we might regularize $X_{+}$ with appropriate gamma priors; however, this approach would degrade the performance of the sparsity model.

On the other hand, when $X_{+}$ is a compound Poisson-ZTP random variable with $\lambda=7$, extreme sparsity levels have virtually no effect on the distribution of the response (Fig~\ref{fig:xpp_pmf_7}). Using the HCPF, we are free to choose any additive distribution for the response model. For discrete response data, we might opt for degenerate, Poisson, binomial, negative binomial or ZTP and for continuous response data, we might select gamma, inverse Gaussian or normal distribution. Furthermore, HCPF explicitly encodes a relationship between non-absence, $Pr(X_{+}\neq 0)$, and the expected non-missing response value, $E[X_{++}]$ (Eq~\ref{eq:sparsity} and Eq~\ref{eq:xpp_mean}), which is defined via the choice of response model. HPF, as a degenerate HCPF model, defines this relationship as $X\sim \delta_{1}$ and $E[X_{++}]=1$, which leads to the poor behavior outside of binary sparse matrices. Along with the flexibility of choosing the most natural element distribution, HCPF is capable of decoupling the sparsity and response models while still encoding a data-specific relationship between the sparsity model and the values of the non-zero responses in expectation.

\section{Hierarchical Compound Poisson Factorization (HCPF)}
\label{sect:hcpf}
Next, we describe the generative process for HCPF and the Gamma-Poisson structure. We explain the intuition behind the choices of the long-tailed Gamma priors. We then present the stochastic variational inference (SVI) algorithm for HCPF.

We can write the generative model of the HCPF with element distribution $p_{\Psi}(x;\theta,\kappa)$ with fixed hyperparameters $\theta$ and $\kappa$ as follows, where $C_{U}$ and $C_{I}$ are the number of users and items, respectively:
\begin{itemize}
\item For each user $u = 1,\dots,C_{U}$
\begin{enumerate}
\item Sample $r_{u} \sim Ga(\rho,\rho / \varrho)$
\item For each component $k$, sample $s_{uk} \sim Ga(\eta,r_{u})$
\end{enumerate}
\item For each item $i= 1,\dots,C_{I}$
\begin{enumerate}
\item Sample $w_{i} \sim Ga(\omega,\omega / \varpi)$
\item For each component $k$, sample $v_{ik} \sim Ga(\zeta,w_{i})$
\end{enumerate}
\item For each user $u$ and item $i$
\begin{enumerate}
\item Sample count $n_{ui} \sim Po(\sum_{k}s_{uk}v_{ik})$
\item Sample response $y_{ui} \sim p_{\Psi}(\theta,n_{ui} \kappa)$
\end{enumerate}
\end{itemize}

\begin{algorithm}[t!]
   \caption{SVI for HCPF}
   \label{alg:svi}
\begin{algorithmic}
   \STATE {\bfseries Initialize:} Hyperparameters $\eta,\zeta,\rho,\varrho,\omega,\varpi,\theta,\kappa,\tau,\xi$ and parameters $t_{u}=\tau,\;t_{i}=\tau,\;b^{r}_{u}=\rho / \varrho,\;a^{s}_{uk}=\eta$
   \STATE$b^{s}_{uk}=\varrho,\;b^{w}_{i}=\omega / \varpi,\;a^{v}_{ik}=\zeta,\;b^{v}_{ik}=\varpi$
   \STATE {\bfseries Fix:}
    \begin{align*}
    a^{r}_{u} &= \rho + K\eta  && a^{w}_{i} = \omega + K\zeta
    \end{align*}
   \REPEAT
   \STATE Sample an observation $y_{ui}$ uniformly from the data set
   \STATE Compute local variational parameters
    \begin{align*}
    \Lambda_{ui} &= \sum_{k} \frac{a^{s}_{uk} a^{v}_{ik}}{b^{s}_{uk} b^{v}_{ik}}\\
    q(n_{ui} = n) &\propto \exp\left \{-\kappa n \Psi(\theta) \right \}h(y_{ui},n\kappa)\frac{\Lambda_{ui}^{n}}{n!}\\
    \varphi_{uik} &\propto \exp \left \{ \Psi(a^{s}_{uk}) - \log b^{s}_{uk}\right.\\
     &\left . \;\;\;\;\;\;\;\;\;+\Psi(a^{v}_{ik}) - \log b^{v}_{ik} \right \}
    \end{align*}
   \STATE Update global variational parameters
    \begin{align*}
    b^{r}_{u} &= (1-t_{u}^{-\xi})b^{r}_{u} + t_{u}^{-\xi}\left (\frac{\rho}{\varrho} + \sum_{k} \frac{a^{s}_{uk}}{b^{s}_{uk}}\right )\\
    a^{s}_{uk} &= (1-t_{u}^{-\xi})a^{s}_{uk} + t_{u}^{-\xi}\left (\eta + C_{I} E[n_{ui}] \varphi_{uik}\right )\\
    b^{s}_{uk} &= (1-t_{u}^{-\xi})b^{s}_{uk} + t_{u}^{-\xi}\left (\frac{a^{r}_{u}}{b^{r}_{u}} + C_{I} \frac{a^{v}_{ik}}{b^{v}_{ik}}\right )\\
    b^{w}_{i} &= (1-t_{i}^{-\xi})b^{w}_{i} + t_{i}^{-\xi}\left (\frac{\omega}{\varpi} + \sum_{k} \frac{a^{v}_{ik}}{b^{v}_{ik}}\right )\\
    a^{v}_{ik} &= (1-t_{i}^{-\xi})a^{v}_{ik} + t_{i}^{-\xi}\left (\zeta + C_{U} E[n_{ui}] \varphi_{uik}\right )\\
    b^{v}_{ik} &= (1-t_{i}^{-\xi})b^{v}_{ik} + t_{i}^{-\xi}\left (\frac{a^{w}_{i}}{b^{w}_{i}} + C_{U} \frac{a^{s}_{uk}}{b^{s}_{uk}}\right )
    \end{align*}
   \STATE Update learning rates
    \begin{align*}
    t_{u} &= t_{u} + 1    &&t_{i} = t_{i} + 1
    \end{align*}
   \STATE (Optional) Update hyperparameters $\theta$ and $\kappa$
   \UNTIL{Validation log likelihood converges}
\end{algorithmic}
\end{algorithm}
The mean field variational distribution for HCPF is given by
\begin{align*}
q(r_{u}\mid a^{r}_{u}, b^{r}_{u})
q(s_{uk}\mid a^{s}_{uk},b^{s}_{uk})
q(w_{i}\mid a^{w}_{i},b^{w}_{i})\\
q(v_{ik}\mid a^{v}_{ik},b^{v}_{ik})
q(z_{ui}\mid \boldsymbol{\varphi_{ui}})
q(n_{ui}).
\end{align*}

The choice of long tail gamma priors has substantial implications for the response model in a collaborative filtering framework.The effect of the gamma prior on a particular user's responses is to effectively characterize her average response. Similarly, a gamma prior on a particular item models the average users' response for that item. The long tail gamma prior assumption for users allows some users to have unusually high responses. For instance, in the donations data, we might expect to observe a few donors who make extraordinarily large donations to a few projects. This is not appropriate for movie ratings, since the maximum rating is $5$, and a substantial number of non-missing ratings are fives. The long tail gamma prior for items allow a few items to receive unusually high average responses. This is a useful property of all of our data sets: we imagine that a few projects may attract particularly large donations, a few blogs may receive a lot of \emph{likes}, or a few movies receive an unusually high average rating.

The choice of long tail gamma priors has different implications in terms of the sparsity model. The gamma prior on a particular user models how active she is, that is, how many items she has responses for. The long tail assumption on the sparsity model implies that there are unusually active users (e.g., cinephiles or frequent donors). The long tail assumption for items corresponds to very popular items with a large number of responses. Note that the movies with the most ratings are not necessarily the highest rated movies.

We leverage the fact that the contributions of Poisson factors can be written as a multinomial distribution~\cite{cemgil2009bayesian}. Using this, we can write out the stochastic variational inference algorithm for HCPF~\cite{hoffman2013stochastic}, where $\tau$ and $\xi$ are the learning rate delay and learning rate power, and $\tau > 0$ and $0.5<\xi<1.0$ (Alg~\ref{alg:svi}).  Note that the HCPF is not a conjugate model due to $q(n_{ui})$. For other variational updates, we only need the statistics $E[n_{ui}]$. For that purpose, we calculate $q(n_{ui} = n)$ explicitly for $n=1,\dots,N_{tr}$. The choice for the truncation value, $N_{tr}$, depends on $\theta, \kappa$, and $\Lambda_{ui}$. We set $N_{tr}$ using the expected range of $\Lambda_{ui}$ and $y_{ui}$ as well as fixed $\theta$ and $\kappa$.

HCPF reduces to HPF when we set $q(n_{ui}) = \delta_{y_{ui}}$. The specific form of $q(n_{ui})$ for different additive EDMs is given in Table~\ref{table:edm}.

\section{Results}
\label{sect:results}
\subsection{Data sets for collaborative filtering}
We performed matrix factorization on $12$ different data sets with different levels of sparsity, response characteristics, and sizes (Table~\ref{table:datasets}). The rating data sets include \textit{amazon} fine food ratings~\cite{mcauley2013amateurs}, \textit{movielens}~\cite{harper2015movielens}, \textit{netflix}~\cite{bell2007lessons} and \textit{yelp}, where the responses are a star rating from $1$ to $5$. The only exception is \textit{movielens} where the maximum rating is $10$. The social media data sets include \textit{wordpress} and \textit{tencent}~\cite{niu2012tencent}, where the response is the number of likes, a non-negative integer value. Commercial data sets include \textit{bestbuy}, where the response is the number of user visit to a product page. The biochemistry data sets include \textit{merck}~\cite{ma2015deep}, which captures molecules (users) and chemical characteristics (items) where the response is the chemical activity. In \textit{echonest}~\cite{bertin2011million}, the response is the number of times a user listened a song. The donation data sets \textit{donation} and \textit{donorschoose} includes donors and projects, where the response is the total amount of a donation in US dollars. The genomics data set \textit{geuvadis} includes genes and individuals, where the response is the gene expression level for a user of a gene~\cite{lappalainen2013}. Both \textit{bestbuy} and \textit{merck} are nearly binary matrices, meaning that the vast majority of the non-missing entries are one. On the other hand, \textit{donation}, \textit{donorschoose}, and \textit{geuvadis} have a continuous response variable.

\subsection{Experimental setup}
We held out $20\%$ and $1\%$ of the non-missing entries for testing ($\mathcal{Y}_{NM}^{test}$) and validation, respectively. We also sampled an equal number of missing entries for testing ($\mathcal{Y}_{M}^{test}$) and validation. When calculating test and validation log likelihood, the log likelihood of the missing entries is adjusted to reflect the true sparsity ratio. Test log likelihood of the missing ($\mathcal{L}_{M}$) and non-missing entries ($\mathcal{L}_{NM}$) as well as the test log likelihood of a non-missing entry conditioned on that it is not missing ($\mathcal{L}_{CNM}$) are calculated as
\begin{align*}
\mathcal{L}_{M} &= \sum_{\mathcal{Y}_{M}^{test}}\log Po(n=0 \mid \Lambda_{ui})\\
\mathcal{L}_{NM} &= \sum_{\mathcal{Y}_{NM}^{test}}\log \sum_{n=0}^{N_{tr}} p_{\Psi}(y_{ui}^{test};\theta,n\kappa) Po(n\mid \Lambda_{ui})\\
\mathcal{L} &= \frac{0.2 (\# \textit{total missing})}{\mid \mathcal{Y}_{M}^{test} \mid} \mathcal{L}_{M} +  \mathcal{L}_{NM}\\
\mathcal{L}_{CNM} &=\sum_{\mathcal{Y}_{NM}^{test}} \log \sum_{n=1}^{N_{tr}} p_{\Psi}(y_{ui}^{test};\theta,n\kappa) ZTP(n \mid \Lambda_{ui}).
\end{align*}
In HCPF, we fix $K=160$, $\xi = 0.7$ and $\tau = 10,000$ after an empirical study on smaller data sets. To set hyperparameters $\theta$ and $\kappa$, we use the maximum likelihood estimates of the element distribution parameters on the non-missing entries. From the number of non-missing entries in the training data set, we inferred the sparsity level, effectively estimating $E[n_{ui}]$ empirically (note that we assume $E[n_{ui}]$ is the same for every user-item pair). We then used $E[n_{ui}]$ to set the factorization hyperparameters $\eta, \zeta, \rho, \varrho, \omega, \varpi$. To create heavy tails and uninformative gamma priors, we set $\varpi=\varrho=0.1$ and $\omega=\rho=0.01$. We then assumed that the contribution of each factor is equal, and set $\eta = \varrho \sqrt{E[n_{ui}]/K}$ and $\zeta = \varpi \sqrt{E[n_{ui}]/K}$. When training on the non-missing entries only, we simply assume that the sparsity level is very low ($0.001$), and from that we set the parameters as usual, and divide the maximum likelihood estimate of $\kappa$ by $E[n_{ui}]$. Earlier work noted that HPF is not sensitive to hyperparameter settings within reason~\cite{gopalan2013scalable}.

\begin{table}[t!]
\caption{{\bf Data set characteristics} Number of rows, columns, non-missing entries, and the ratio of missing entries to the total number of entries (sparsity) for the data sets we analyzed.}
\label{table:datasets}
\vskip -0.15in
\begin{center}
\begin{scriptsize}
\begin{sc}
\begin{tabular}{lrrrr}
\hline
{data set} &     \# rows &    \# cols &  sparsity & \# non-missing \\
\hline
\abovespace
amazon       &   256,059 &   74,258 &  0.999970 &       568,454 \\
movielens    &   138,493 &   26,744 &  0.994600 &    20,000,263 \\
netflix      &   480,189 &   17,770 &  0.988224 &   100,483,024 \\
yelp         &   552,339 &   77,079 &  0.999947 &     2,225,213 \\
wordpress    &    86,661 &   78,754 &  0.999915 &       581,508 \\
tencent      & 1,358,842 &  878,708 &  0.999991 &    10,618,584 \\
bestbuy      & 1,268,702 &   69,858 &  0.999979 &     1,862,782 \\
merck        &   152,935 &   10,883 &  0.970524 &    49,059,340 \\
echonest     & 1,019,318 &  384,546 &  0.999877 &    48,373,586 \\
donation     &   394,266 &      82  &  0.965831 &     1,104,687 \\
dchoose      & 1,282,062 &  525,019 &  0.999996 &     2,661,820 \\
\belowspace
geuvadis     &     9,358 &      462 &  0.462122 &     2,325,461 \\
\hline
\end{tabular}
\end{sc}
\end{scriptsize}
\end{center}
\vskip -0.35in
\end{table}
To identify the best response model for HCPF, we ran SVI with all seven additive EDM distributions in Table~\ref{table:edm}. We first fit all HCPF models by sampling from the full matrix. We calculated $\mathcal{L}, \mathcal{L}_{M}, \mathcal{L}_{NM}$ and $\mathcal{L}_{CNM}$. In Section~\ref{sect:tll}, we compare HCPF and HPF in $\mathcal{L}$. In the second analysis, we only used the non-missing entries for training and calculated $\mathcal{L}_{NM}$. In Section~\ref{sect:nz_tll}, we compared $\mathcal{L}_{CNM}$ of the first analysis to $\mathcal{L}_{NM}$ of the second analysis.

\begin{table*}[t]
\caption{{\bf Test log likelihood.} Per-thousand-entry test log likelihood for HCPF and HPF trained on the full matrix for twelve data sets. Discrete HCPF models and HPF are not applicable to continuous data sets (N/A). Element distribution acronyms in Table~\ref{table:edm}.}
\label{table:tll}
\begin{center}
\begin{scriptsize}
\begin{sc}
\begin{tabular}{lrrrrrrrrrrrr}
\hline
\abovespace
\belowspace
         & \rot{amazon} & \rot{movielens} & \rot{netflix} &   \rot{yelp} & \rot{wordpress} & \rot{tencent} & \rot{echonest} & \rot{bestbuy} &  \rot{merck}  & \rot{donation} & \rot{dchoose}   & \rot{geuvadis} \\
\hline
\abovespace
HCPF-N   &  -0.388      &  -27.480        &  -50.461      &  \bf-0.625   &  -0.909         &  -0.148       &  -1.502        &  -0.231       &  -129.225     &  -284.178      &    -0.100       &  -4424.288     \\
\rowcolor{lightgray}
HCPF-GA  &  -0.402      &  -28.546        &  -51.558      &  -0.648      &  \bf-0.804      &  \bf-0.129    &  -1.697        &  -0.221       &  -107.569     &  -260.704      &    \bf-0.095    &  \bf-2527.897  \\
HCPF-IG  &  -0.401      &  -27.764        &  -51.152      &  -0.657      &  -0.876         &  \bf-0.129    &  -1.648        &  \bf-0.215    &  -93.989      &  \bf-254.055   &    \bf-0.095    &  -36327.644    \\
\rowcolor{lightgray}
HCPF-PO  &  -0.392      &  \bf-27.056     &  -51.332      &  -0.636      &  -0.872         &  -0.183       &  \bf-1.447     &  -0.298       &  -136.609     &        N/A     &          N/A    &        N/A     \\
HCPF-BI  &  \bf-0.387   &  -27.781        &  \bf-50.152   &  -0.708      &  -0.930         &  -0.174       &  -1.871        &  -0.289       &  -116.886     &        N/A     &          N/A    &        N/A     \\
\rowcolor{lightgray}
HCPF-NB  &  -0.404      &  -28.886        &  -55.201      &  -0.701      &  -0.818         &  -0.143       &  -1.709        &  -0.342       &  -111.613     &        N/A     &          N/A    &        N/A     \\
HCPF-ZTP &  -0.389      &  -27.932        &  -52.698      &  -0.661      &  -0.850         &  -0.170       &  -1.846        &  -0.304       &  -113.856     &        N/A     &          N/A    &        N/A     \\
\rowcolor{lightgray} \belowspace
HPF      &  -1.453      &  -76.182        &  -88.490      &  -2.026      &  -1.378         &  -0.583       &  -3.246        &  -0.289       &  \bf-86.492   &        N/A     &          N/A    &        N/A     \\
\hline
\end{tabular}
\end{sc}
\end{scriptsize}
\end{center}
\vskip -0.10in
\end{table*}

\begin{table*}[t]
\caption{{\bf Non-missing test log likelihood.} Per non-missing entry test log likelihood of HCPF and HPF trained on the full matrix and on the non-missing entries only (marked as `F' and `NM,' respectively). When trained on the full matrix, the conditional non-missing test log likelihood is reported.}
\label{table:nz_tll}
\begin{center}
\begin{scriptsize}
\begin{sc}
\begin{tabular}{llrrrrrrrrrrrr}
\hline
\abovespace\belowspace
{} & {}      & \rot{amazon} & \rot{movielens} & \rot{netflix} &   \rot{yelp} & \rot{wordpress} & \rot{tencent} & \rot{echonest} & \rot{bestbuy} &  \rot{merck}  & \rot{donation} & \rot{dchoose} & \rot{geuvadis} \\
\hline
\abovespace
HCPF-N  & F  &   -1.692 &    -2.290 &    -1.642 &    -1.734 &    -3.026 &    -4.391 &    -3.201 &     -0.872 &    -2.855 &    -4.985 &       -6.806 &    -6.841 \\
        & NM &   -6.302 &    -2.268 &    -1.617 &    -3.573 &    -2.860 &    -3.933 &    -2.706 &    -16.754 &    -2.403 &    -4.881 &       -7.421 &    -7.173 \\
\rowcolor{lightgray}
HCPF-GA & F  &   -1.909 &    -2.417 &    -1.708 &    -1.876 &    -1.754 &    -2.500 &    -2.049 &     -0.835 &    -2.151 &    -4.338 &       -5.387 & \bf-3.341 \\
\rowcolor{lightgray}
        & NM &   -4.102 &    -2.310 &    -1.649 &    -2.776 &    -2.234 &    -2.761 &    -2.053 &    -17.265 &    -2.079 &    -4.213 &       -6.906 &    -4.129 \\
HCPF-IG & F  &   -2.077 &    -2.363 &    -1.665 &    -1.999 & \bf-1.440 & \bf-2.040 & \bf-1.752 &     -0.782 &    -1.685 & \bf-4.132 &    \bf-5.340 &   -66.228 \\
        & NM &   -5.965 &    -2.690 &    -3.419 &    -5.589 &    -7.017 &    -7.470 &    -6.979 &     -8.247 &    -7.474 &    -7.135 &       -6.155 &   -10.121 \\
\rowcolor{lightgray}
HCPF-PO & F  &   -1.877 &    -2.255 &    -1.756 &    -1.856 &    -2.448 &    -7.960 &    -3.017 &     -1.001 &    -3.186 &       N/A &          N/A &      N/A \\
\rowcolor{lightgray}
        & NM &   -7.206 &    -9.819 &    -5.608 &    -6.210 &    -5.234 &   -12.633 &    -6.145 &     -3.175 &    -6.673 &       N/A &          N/A &      N/A \\
HCPF-BI & F  &\bf-1.559 &    -2.208 & \bf-1.586 & \bf-1.704 &    -2.449 &    -6.911 &    -3.085 &     -0.694 &    -2.457 &       N/A &          N/A &      N/A \\
        & NM &   -2.113 &    -2.214 &    -1.626 &    -1.841 &    -2.649 &    -5.480 &    -2.448 &     -1.102 &    -1.993 &       N/A &          N/A &      N/A \\
\rowcolor{lightgray}
HCPF-NB & F  &   -2.113 &    -2.510 &    -2.078 &    -2.067 &    -1.958 &    -3.441 &    -2.272 &     -1.349 &    -2.267 &       N/A &          N/A &      N/A \\
\rowcolor{lightgray}
        & NM &   -3.772 &    -2.526 &    -2.082 &    -2.793 &    -2.344 &    -3.272 &    -2.178 &     -3.104 &    -2.085 &       N/A &          N/A &      N/A \\
HCPF-ZTP& F  &   -1.865 &    -2.356 &    -1.821 &    -1.832 &    -2.283 &    -6.924 &    -3.008 &  \bf-0.010 &    -2.323 &       N/A &          N/A &      N/A \\
        & NM &   -6.853 &    -5.324 &    -5.541 &    -5.741 &    -5.406 &   -12.123 &    -5.952 &     -3.002 &    -6.653 &       N/A &          N/A &       N/A \\
\rowcolor{lightgray}
HPF     & F  &  -35.666 &    -9.703 &    -4.623 &   -27.025 &    -7.167 &   -63.794 &   -22.895 &     -0.017 & \bf-1.535 &       N/A &          N/A &      N/A \\
\rowcolor{lightgray} \belowspace
        & NM &   -1.868 & \bf-2.095 &    -1.672 &    -1.940 &    -2.522 &    -6.784 &    -2.640 &     -3.305 &    -1.563 &       N/A &          N/A &      N/A \\
\hline
\end{tabular}
\end{sc}
\end{scriptsize}
\end{center}
\vskip -0.10in
\end{table*}

\subsection{Overall performance}
\label{sect:tll}
In this analysis we quantify how well these models capture both the sparsity and response behavior in ultra sparse matrices. In a movie ratings data set, the question becomes  `Can we predict if a user would rate a given movie and if she does what rating she would give?'. We report the test log likelihood of all twelve data sets. We fit HCPF with normal, gamma, and inverse Gaussian as element distributions for all the data sets. In discrete data sets, we additionally fit HPF and HCPF with Poisson, binomial, negative binomial, and zero truncated Poisson element distributions.

In all ratings data sets (\textit{amazon, movielens, netflix}, and \textit{yelp}), HCPF significantly outperforms HPF (Table~\ref{table:tll}). The relative performance difference is even more pronounced in sparser data sets (\textit{amazon} and \textit{yelp}). When we break down the test log likelihood into missing and non-missing parts, we see that, in sparser data sets, the relative performance of HPF for non-missing entries is much weaker than it is for less sparse data sets. This is expected, as response coupling in HPF is stronger in sparser data sets, forcing the response variables to zero. The opposite is true for HCPF: its performance improves with increasing data sparsity.

In social media activity data (\textit{wordpress} and \textit{tencent}) and the music data set (\textit{echonest}), HCPF shows a significant improvement over HPF (Table~\ref{table:tll}). Unlike the ratings data sets, we have an unbounded response variable with an exponentially decaying characteristic. At first HPF might seem to be a good model for such data; however, the sparsity level is so high that the non-missing point mass concentrates at $1$. This is best seen in the comparison of HPF with \textit{HCPF-ZTP}. Although the Poisson distribution seems to capture response characteristic effectively, the test performance degrades when zero is included as part of the response model (as in HPF).

In the \textit{bestbuy} and \textit{merck} data sets, where the response is near binary, HPF and HCPF performances are very similar. This confirms the observation that HPF is a sufficiently good model for sparse binary data sets. In the financial data sets (\textit{donation} and \textit{donors-choose}), we see that the gamma and inverse Gaussian are better distributional choices than the normal as the element distribution.

\subsection{Response model evaluated using test log likelihood}
\label{sect:nz_tll}
In this section, we investigate which model captures the response most accurately. In a movie ratings data set, the question becomes  `Can we predict what rating a user would give to a movie given that we know she rated that movie?'. In Table~\ref{table:nz_tll}, we report the conditional non-missing test log likelihood of models trained on the full matrix and test log likelihood of the models trained only on the non-missing entries.

First, we note that training HPF only on the non-missing entries results in a better response model than training HPF on the full matrix. The only exception is \textit{bestbuy} where the conditional non-missing test log likelihood is near perfect. This is due to the near binary structure of the data set. When we know if an entry is not missing, then we are fairly confident that it has a value of $1$.

\begin{table*}[t!]
\caption{{\bf Test AUC} Test AUC values for HCPF trained on the full matrix and HPF trained on the binarized full matrix. }
\label{table:auc}
\begin{center}
\begin{scriptsize}
\begin{sc}
\begin{tabular}{llrrrrrrrrrrrr}
\hline
\abovespace\belowspace
{} & {} & \rot{amazon} & \rot{movielens} & \rot{netflix} &   \rot{yelp} & \rot{wordpress} & \rot{tencent} & \rot{echonest} & \rot{bestbuy} &  \rot{merck}  & \rot{donation} & \rot{dchoose} & \rot{geuvadis} \\
\hline
\abovespace
HCPF-N   &        F &    0.8020 &     0.9854 &     0.9720 & \bf0.8768 &     0.8991 &     0.9167 &     0.9111 &     0.8892 &     0.9872 &  \bf0.8804 &      0.6002 &     0.7749 \\
\rowcolor{lightgray}
HCPF-GA  &        F &    0.7986 &     0.9839 &     0.9723 &    0.8651 &     0.8943 &     0.9175 &     0.8912 &  \bf0.8893 &     0.9880 &     0.8775 &      0.6543 &     0.7740 \\
HCPF-IG  &        F &    0.8017 &     0.9856 &     0.9722 &    0.8681 &     0.8953 &     0.9136 &     0.8892 &     0.8830 &     0.9880 &     0.8773 &      0.5967 &     0.7733 \\
\rowcolor{lightgray}
HCPF-PO  &        F & \bf0.8034 &  \bf0.9860 &     0.9721 &    0.8731 &     0.8989 &     0.9110 &  \bf0.9124 &     0.8859 &     0.9884 &         N/A &          N/A &         N/A \\
HCPF-BI  &        F &    0.8004 &     0.9848 &     0.9725 &    0.8452 &     0.8985 &     0.9164 &     0.8932 &     0.8878 &  \bf0.9885 &         N/A &          N/A &         N/A \\
\rowcolor{lightgray}
HCPF-NB  &        F &    0.7989 &     0.9844 &  \bf0.9729 &    0.8510 &  \bf0.9002 &     0.9119 &     0.8908 &     0.8168 &     0.9877 &         N/A &          N/A &         N/A \\
HCPF-ZTP &        F &    0.8025 &     0.9849 &     0.9717 &    0.8640 &     0.8999 &  \bf0.9176 &     0.8922 &     0.8440 &     0.9876 &         N/A &          N/A &         N/A \\
\rowcolor{lightgray} \belowspace
HPF      &        B &    0.8021 &     0.9855 &     0.9718 &    0.8508 &     0.8971 &     0.9146 &     0.8901 &     0.8860 &     0.9879 &     0.8780 &   \bf0.6552 &  \bf0.7769 \\
\hline
\end{tabular}
\end{sc}
\end{scriptsize}
\end{center}
\vskip -0.10in
\end{table*}

Secondly, we investigate if modeling the missing entries explicitly helps the response model. We compare HPF trained on the non-missing entries to HCPF trained on the full matrix. HCPF-BI outperforms HPF in all ratings data sets except for \textit{movielens}. A similar pattern can be seen in social media and music data sets where HCPF-IG seems to be the best model.

This phenomenon can be attributed to better identification of the relationship between the sparsity model and the response model. Although HPF trained on the full matrix can also capture this relationship, the high sparsity levels force HPF to fit near-zero Poisson parameters, hurting the prediction for response. In ratings data sets, HCPF captures the notion that more likely people are to consume an item, higher their responses are. In \textit{movielens} and \textit{netflix} for instance, we know that the most watched movies tend to have higher ratings. In social media data sets, the relation between the reach and popularity is captured. The more followers a blog has, the more content it is likely to produce, and the more reaction it will get. A similar correlation exists for the users as well. The more active users are also the more responsive ones. In summary, HCPF can capture the relationship between the non-missingness of an entry and the actual value of the entry if it is non-missing. Any NMF algorithm that makes the missing at random assumption would miss this relationship.

One might argue that perhaps HPF is not a good model because the underlying response distribution is not Possion like. Comparing the rows marked `NM', we observe that there is some truth to this argument. In \textit{netflix} and \textit{yelp} data sets, \textit{HCPF-BI} outperforms HPF; however, we also see that training \textit{HCPF-BI} on the full matrix is even better. A similar argument can be made for \textit{HCPF-GA} in social media data sets (\textit{wordpress} and \textit{tencent}). The flexibility of HCPF is a key factor to identifying the underlying response distribution. The ability to model sparsity and response at the same time gives HCPF a further edge in modeling response.

\subsection{Sparsity model evaluated using AUC}
\label{sect:auc}
In a movie ratings data set or a purchasing data set, one important question is  `Can we predict if a user would rate a given movie or buy a certain item?'.  To understand the quality of our performance on this task, we evaluated the sparsity model separately by computing the area under the ROC curve (AUC), fitting all HCPF models by sampling from the full matrix and HPF to the binarized full matrix (Table~\ref{table:auc}). To calculate the AUC, the true label is whether the data are missing ($0$) or non-missing ($1$), and we used the estimated probability of a non-missing entry, $Pr(X_{+}\neq 0)$, as the model prediction. As discussed in Section~\ref{sect:tll}, when we fit HPF to the full matrix, we compromise performance on sparsity and response. HCPF, on the other hand, enjoys the decoupling effect while preserving the relationship in expectation (see Eq~\ref{eq:xpp_mean}). In $10$ of the $12$ data sets, we get an improvement over HPF (Table~\ref{table:auc}); this is somewhat surprising as the HPF is specialized to this task; this illustrates the benefit of coupling the sparsity and response models in expectation. Better modeling of the response would naturally lead to a better sparsity model.

\section{Conclusion}
In this paper, we first proved that a zero truncated compound Poisson distribution converges to its element distribution as sparsity increases. The implication of this theorem for HPF is that the non-missing response distribution concentrates at $1$, which is not a good response model unless we have a binary data set. Inspired by the convergence theorem, we introduce HCPF. Similar to HPF, HCPF has the favorable Gamma-Poisson structure to model long-tailed user and item activity. Unlike HPF, HCPF is capable of modeling binary, non-negative discrete, non-negative continuous and zero-inflated continuous data. More importantly, HCPF decouples the sparsity and response models, allowing us to specify the most suitable distribution for the non-missing response entries. We show that this decoupling effect improves the test log likelihood dramatically when compared to HPF on high-dimensional, extremely sparse matrices. HCPF also shows superior performance to HPF trained exclusively on non-missing entries in terms of modeling response. Finally, we show that HCPF is a better sparsity model than HPF, despite HPF targeting this sparsity behavior.

For future directions, we will investigate the implications of the decoupling theorem in other Bayesian settings. We will also explore hierarchical latent structures for the element distribution.
\newpage
\section*{Acknowledgements}
BEE was funded by NIH R00 HG006265, NIH R01 MH101822, and a Sloan Faculty Fellowship. MEB was funded in part by the Princeton Innovation J. Insley Blair Pyne Fund Award. We would like to thank Robert E. Schapire for valuable discussions.
\bibliography{compound_poisson_factorization}
\bibliographystyle{icml2016}

\newpage
\twocolumn[
\section*{Appendix}
\subsection*{Proof of Theorem 2}
\begin{proof}
Follows directly from Theorem 1.
\end{proof}

\subsection*{Proof of Remark 1}
\begin{proof}
Let $M_{X}(t)$ be the moment generating function (MGF) of $X\sim p_{\Psi}(x;\theta,\kappa)$, then MGF of $X_{+}$ is given by $M_{X_{+}}(t) = e^{\Lambda (M_{X}(t)-1)}$. MGF of an ordinary Poisson random variable $Y$ with parameter $\Lambda$ is $M_{Y}(t) = e^{\Lambda (e^{t}-1)}$. If $M_{X_{+}}(t) = M_{Y}(t)$, then $M_{X}(t) = e^{t}$ which is the MGF for degenerate distribution $\delta_{1}$. If $M_{X}(t) = e^{t}$, then $M_{X_{+}}(t) = M_{Y}(t)$.
\end{proof}

\subsection*{Proof of Theorem 3}
\begin{proof}
Let $M_{X}(t)$ be the MGF of $X\sim p_{\Psi}(x;\theta,\kappa)$ and $\left \{X_{++}^{m}\right \}_{m=1}^{\infty}$ be a sequence of random variables where $X_{++}^{m} = X_{+}^{m} \mid X_{+}^{m} \neq 0$ and $X_{+}^{m} \sim p_{\Psi}(x;\theta,\kappa,\Lambda=\frac{1}{m})$. The MGF of $X_{++}^{m}$ is given by
\begin{align*}
M_{X_{++}^{m}}(t) &= \frac{e^{M_{X}(t)/m}-1}{e^{1/m}-1}.
\end{align*}
Since $\lim_{m \rightarrow \infty}M_{X_{++}^{m}}(t) = M_{X}(t)$, $X_{++}$ converges to $X$ in distribution as $\Lambda$ goes to zero.
\end{proof}
]

\end{document}